\documentclass[10pt,letterpaper]{article}
\usepackage{cogsci}
\cogscifinalcopy 
\usepackage[
    backend=biber,
    style=apa,
    natbib=true,
    doi=false,
    isbn=false,
    url=false,
]{biblatex}
\usepackage{graphicx}
\usepackage{pslatex}
\usepackage{float}
\usepackage{hyperref}
\usepackage{subcaption}
\usepackage{amsmath}
\usepackage{amsfonts}
\usepackage{pifont}
\usepackage{booktabs}
\usepackage[hang,flushmargin]{footmisc}

\renewcommand{\footnotesize}{\fontsize{8}{10}\selectfont}

\hypersetup{
    colorlinks,
    citecolor=black,
    filecolor=black,
    linkcolor=black,
    urlcolor=black
}

\title{Effective Explanations Support Planning Under Uncertainty}
 
\author{
\textbf{Hanqi Zhou}$^{*,1,2,3,4}$, \textbf{Britt Besch}$^{5}$, \textbf{Charley M. Wu}$^{2,3,4}$ \& \textbf{Tobias Gerstenberg}$^{6}$ \\
$^*$\texttt{hanqi.zhou@uni-tuebingen.de}\\
$^1$University of T\"ubingen $^2$Technical University Darmstadt $^3$Hessian.AI \\ $^4$Max Planck Institute for Biological Cybernetics $^5$University of Cambridge $^6$Stanford University
}

\begin{document}

\maketitle

\begin{abstract}
Explaining how to get from A to B can be challenging. It requires mentally simulating what the listener will do based on what they are told. 
To capture this process, we propose a computational model that converts utterances into action plans: a large language model translates an explanation into program-like guidance (a policy prior and value map), and a planning agent executes it under partial observability. 
We score explanations by the efficiency and reliability of the resulting paths, penalizing replanning.
Across four preregistered experiments, we collect a corpus of 1,200 explanations over 24 maps, elicit helpfulness judgments, measure baseline navigation, and test behavior with explanations of differing quality. 
Higher-scored explanations are judged more helpful and improve navigation: participants with explanations outperform those without, and high-scoring explanations help more than low-scoring ones. 
Together, these results show procedural explanation as utility-guided communication shaped by how language can be grounded into action under uncertainty.
 
\end{abstract}

\section{Introduction}
People routinely explain \emph{how} to do things: how to find a building on a campus, how to assemble furniture, or how to troubleshoot a device. 
Such procedural explanations are rarely exhaustive descriptions of all possible actions. Instead, an effective explanation selects what matters and structures it in a way that enables the listener to act successfully.

Viewed this way, procedural explanations are a special case of explanations aimed at supporting actions rather than mere belief change. Existing work often emphasizes the role of such explanations in learning, by helping listeners infer the underlying causal structure that generates the observed data, allowing them to generalize their knowledge to new situations \citep{lombrozo2009explanation, lombrozo201214, kirfel2022inference, chandra2024cooperative}. 
In many everyday settings, however, the central function is immediate action guidance under constraints: listeners have limited attention and memory, and speakers are uncertain about the context in which the explanation will be used. On this view, explanation quality is not just measured by semantic adequacy or completeness, but also the pragmatic value in helping a listener make the right decisions.

This perspective aligns with broader theories of communication as \emph{utility-guided inference} \citep[see][]{sumers2024reconciling,harding2025communication}. 
In Gricean pragmatics, speakers cooperate by being informative while respecting conversational constraints \citep[e.g., giving just enough information;][]{grice1975logic}. 
Probabilistic approaches model this utterance choice as boundedly rational: speakers trade off expected communicative success against costs like utterance length or complexity \citep{degen2023rational}. 
The Rational Speech Act (RSA) framework instantiates this idea by formalizing language use as probabilistic inference over speakers' and listeners' mental states \citep{frank2012predicting, goodman2016pragmatic, chandra2024cooperative}.
Complementing this approach, computational models of pedagogy show that teaching goals change what speakers choose to convey, and listeners can infer, yielding listener behavior that's distinct from acting or observing \citep{shafto2014rational, austerweil2019learning, chen2024hierarchical, velez2023teachers}. 
Thus, a common theme is that communicative acts are shaped by their downstream consequences for an intended recipient.

Here, we use navigation as a case study. When one person explains how to get from A to B, the explainer knows the environment while the listener must act from the utterance plus local observations \citep{fried2018speaker}. Work on wayfinding shows that speakers adapt to this asymmetry by emphasizing landmarks, decision points, and hierarchical structure rather than every step \citep{schwering2017wayfinding, denis2007perspectives, baltaretu2015improving}---suggesting a normative principle in which effective explanations allocate information to minimize expected downstream mistakes, not to maximize descriptive completeness.

\begin{figure*}[t]
    \centering
    \includegraphics[width=1.0\linewidth]{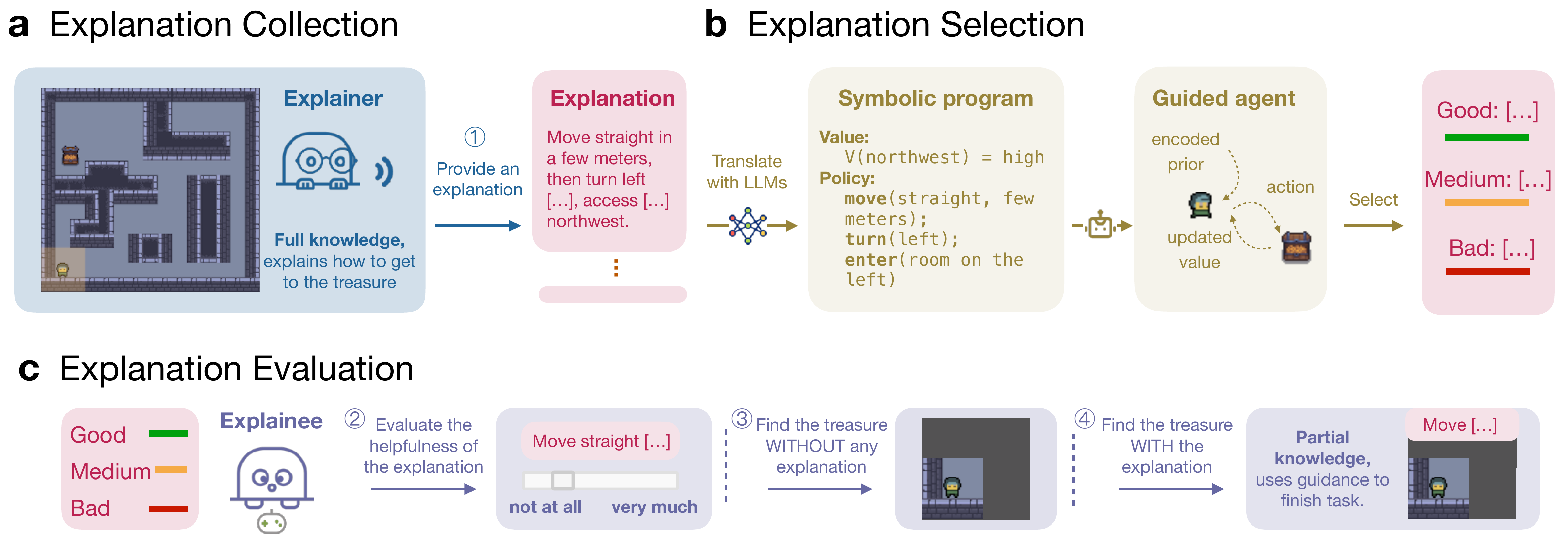}
    \caption{
    \textbf{Modeling language-guided navigation and experimental pipeline.}
    (a) Explanation collection (Exp.~1): An explainer with full knowledge of the environment generates natural-language explanations for an explainee acting under partial observability.
    (b) Explanation modeling and selection: Free-form text explanations are translated by LLMs into symbolic programs and evaluated by a simulated agent that plans and acts based on the symbolic program. The performance of this agent is used to rate the explanation quality.
    (c) Explanation evaluation: Model-ranked explanations are evaluated via behavioral experiments measuring perceived helpfulness (Exp.~2), baseline performance without explanation (Exp.~3), and performance with bad/medium/good explanations (Exp.~4). 
    }
    \label{fig:paradigm}
\end{figure*}

At the same time, navigation is an open-ended instruction problem: environments can be large, and explainers can describe routes in many valid ways. This open-endedness has fueled interest in \emph{machine-generated explanations} from large language models (LLMs), which can produce fluent instructions on the fly \citep{joshi2023machine, saha2023can}. 
Yet fluency is not enough---generated explanations can be causally wrong or emphasize details that do not help a listener act, motivating behavioral evaluation of whether they actually improve performance. This opens an opportunity for cognitive science and AI to study how language is grounded into executable actions under uncertainty \citep{chen2011learning, ahn2022can}, raising a central question: \emph{what makes an explanation effective for guiding action}?

We address this question in a controlled navigation domain that isolates key ingredients of procedural explanation under partial observability.
An \textit{explainer} has global knowledge of the environment and goal, while an \textit{explainee} navigates based on the explanation and their local observations. 
Across four experiments (\autoref{fig:paradigm}), we collect a corpus of human-written navigation explanations (Exp.~1), measure perceived helpfulness (Exp.~2), establish baseline navigation performance without explanations (Exp.~3), and test how explanations affect navigation when available (Exp.~4). We then evaluate a computational account in which explainers trade off expected gains in explainee success against communicative costs. Bringing together what people say (explanations), what they think is helpful (judgments), and what they do (navigation outcomes), we characterize procedural explanation as a resource-rational strategy for acting and communicating under uncertainty and limited shared knowledge.

\section{Experiments}
\begin{figure*}[!ht]
    \centering
    \includegraphics[width=1.0\textwidth]{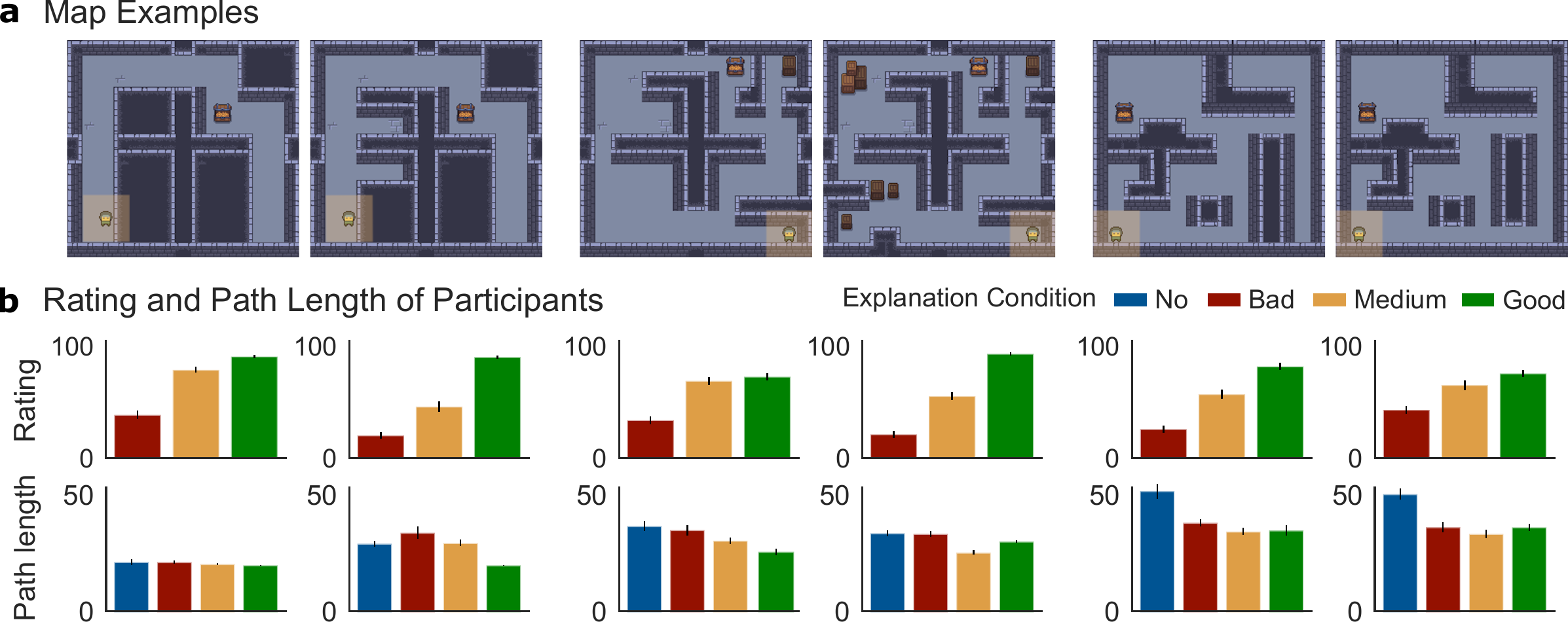}
    \caption{\textbf{Paired map examples and behavioral effects of explanation quality.}
    (a) Example map pairs: matched overall layouts with small local changes (obstacles/structure).
    (b) Path length (top) and helpfulness ratings (bottom) by condition (None, Bad, Medium, Good) for each map pair (columns). Bars show means $\pm$ SE; dots show individual participants. Higher-quality explanations consistently increase perceived helpfulness, but path length gains vary across pairs.}
    \label{fig:map-example}
\end{figure*}

All four experiments were preregistered\footnote{See full materials here:
\href{https://github.com/cicl-stanford/explaining_how_cogsci26}{github.com/cicl-stanford/explaining\_how\_cogsci26}} and conducted on Prolific.
We use a grid-based ``dungeon'' navigation environment with 24 distinct maps, organized into 12 near-matched pairs that differ only by small, localized edits (see \autoref{fig:map-example}a for examples). Each map has a start location and a hidden goal (treasure). An explainer sees the full map and writes a natural-language message to guide a partner to the treasure. The explainee either rates the message's helpfulness or navigates under partial observability, viewing only a local region around their current position.

\noindent \textbf{Exp.~1: Explanation collection.}
To collect a dataset of explanations, we recruited $N=50$ participants to act as explainers (\autoref{fig:paradigm}a \ding{192}). 
On each trial, explainers saw the full map and wrote a free-text message to help a partner find the treasure. They were instructed: ``Please send a message to your partner that will help them find the treasure. Remember that your partner can only see the highlighted area -- they cannot see the whole map.''
Each participant wrote explanations for 24 maps presented in randomized order. We recorded explanation text and response times (average completion $25.64 \pm 11.70$ minutes). We performed minimal preprocessing on the free-form explanations (whitespace normalization; no content edits). 

\noindent \textbf{Exp.~2: Helpfulness judgments.}
To acquire participant ratings about the helpfulness of explanations generated in Exp.~1, we recruited a separate group of $N=50$ participants.
For each map, we selected three explanations (``good'', ``medium'', ``bad'') following a model-based ranking procedure (\autoref{fig:paradigm}c \ding{193}). 
Participants viewed each map and rated how helpful each message was under the instruction ``Please evaluate the following messages by rating how helpful they are for finding the treasure''. Both maps and messages were presented in a randomized order (avg. $22.62 \pm 10.66$ mins). 
We preregistered the hypothesis that participants' helpfulness ratings would reflect the model-based ranking order Good $>$ Medium $>$ Bad.

\noindent \textbf{Exp.~3: Baseline navigation without explanations.}
To provide a baseline measure of map difficulty and between-map variability in path length, we recruited
$N=50$ participants to navigate each map under partial observability with the instruction ``Find the treasure in as few steps as possible''.
The interface revealed a local field-of-view around the current position (\autoref{fig:paradigm}c \ding{194}).
We recorded participants' trajectories and completion time (avg. $10.89 \pm 4.80$ mins). 

\noindent \textbf{Exp.~4: Navigation with explanations.}
To test how explanation quality shapes navigation efficiency, we recruited $N=150$ participants to navigate each map under partial observability (with the same instruction as Exp.~3) but also receiving one explanation per map (\autoref{fig:paradigm}c \ding{195}, avg. $13.46 \pm 5.92$ mins). Explanation quality (Good/Medium/Bad) was manipulated within-subjects using counterbalanced lists so each participant saw equal numbers of maps at each quality level. 
Our preregistered hypothesis was that mean path length would follow the model ranking (Good $<$ Medium $<$ Bad) and that explanations would reduce path length overall (With explanation $<$ No explanation).

\section{Computational model}
An explanation is useful to the extent that it can be translated into guidance that improves downstream navigation. We model navigation as a POMDP parameterized by the fully observed world~$w$ (map, start, goal): the agent takes actions $a_t$ with transitions $P(s_{t+1}\mid s_t, a_t, w)$ over latent states~$s_t$, observing only local~$o_t$ \citep{kaelbling1998planning}.

\noindent \textbf{LLM translation as a stochastic compiler.}
We use a large language model (LLM) to compile an explanation~$e$ into program-like guidance for planning. We compile explanations into a policy prior and value map, $\tau = \big(\hat{\pi}_{\tau}(a\mid s), \hat{V}_\tau(s)\big)$, where $\hat{\pi}_{\tau}$ is a policy prior over actions and $\hat{V}_{\tau}$ is a value map over planner states $s$. The states $s= \phi(o,w)$ are mapped onto observations $o$ in the current world $w$.  
Because LLM decoding is stochastic, we treat compilation as a conditional distribution $p(\tau \mid e, w)$. 
In practice, we approximate this distribution by repeatedly running the same prompting procedure and obtaining $K$ compilations $\tau^{(k)} \sim \mathcal{T}(e,w)$, where $\mathcal{T}(e,w)$ denotes the output of the LLM under this fixed prompt constructed from $(e,w)$ and fixed decoding settings.

Let $\textsc{Plan}(\tau, s)$ denote the agent's action-selection rule given guidance $\tau$ at state $s$ (e.g., using $\hat{\pi}_{\tau}$ or $\hat{V}_{\tau}$ to bias search). At time $t$,
\begin{align}
    a_t \sim \textsc{Plan}(\tau_t, s_t), \qquad s_t = \phi(o_t,w).
\end{align}
If the current compilation fails to yield usable guidance (e.g., invalid, or repeatedly dead-ends), the agent re-queries the translator for a new sample 
$\tau_t \leftarrow \mathcal{T}(e, w) \quad \text{whenever } \textsc{Fail}(\tau_t, o_t)=1$.

\begin{figure*}[t]
    \centering
    \includegraphics[width=1.0\textwidth]{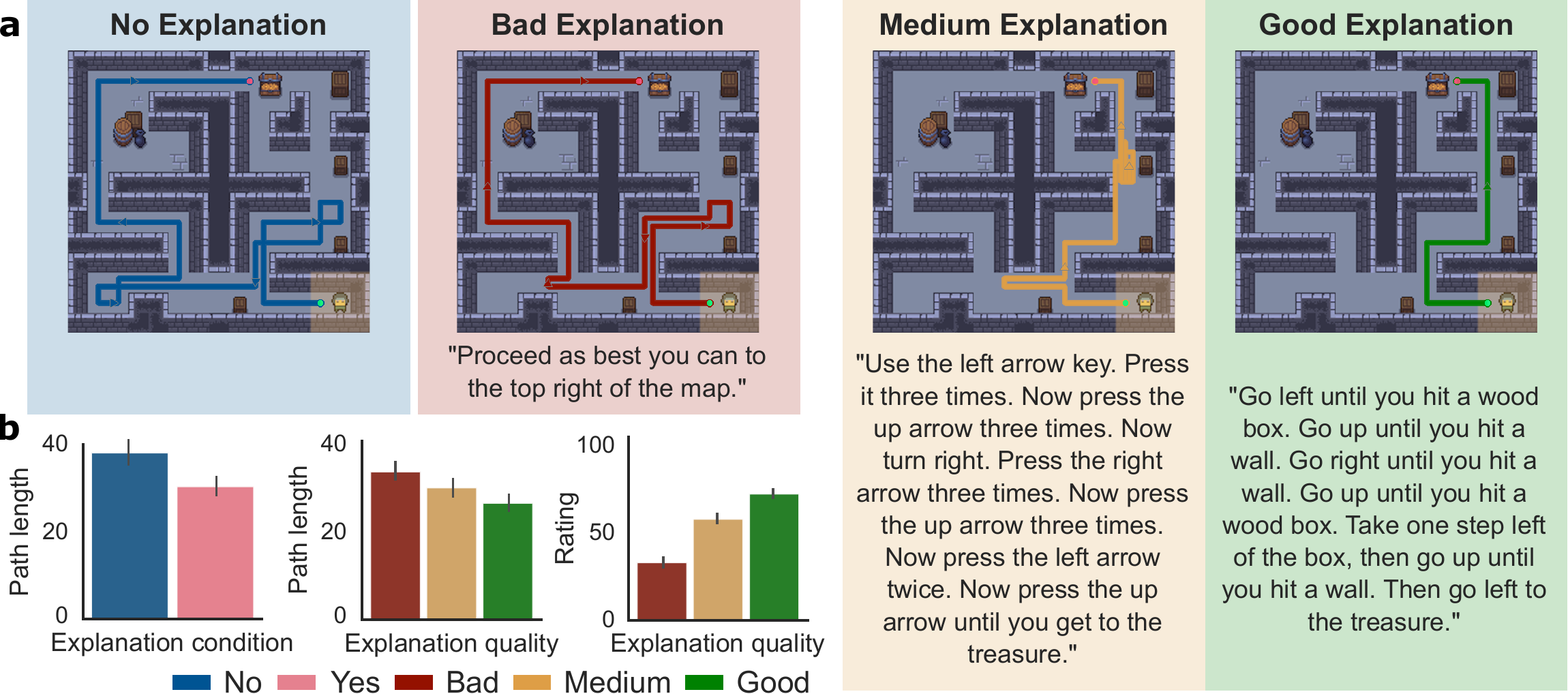}
    \caption{\textbf{Explanations improve navigation efficiency and subjective helpfulness.}
    (a) Example trajectories by condition: no explanation yields inefficient exploration; low-quality explanations are vague procedural; high-quality explanations emphasize relevant landmarks and structure, producing more direct paths.
    (b) Path length by condition (No/Bad/Medium/Good): higher quality yields shorter paths and higher helpfulness ratings.
    Bars show means; black lines bootstrapped 95\% confidence intervals.
    }
    \label{fig:regression}
\end{figure*}

\noindent \textbf{Utility trades off replanning, efficiency, and success.}
We define the utility of an explanation as a function of replanning costs $\textsc{Replan}$ (due to ambiguous explanations), efficiency $\textsc{Len}_{\min}$ (measured by the length of the explanation), and downstream success $\textsc{Succ}$ (whether the goal is reached), measured over $N$ independent attempts:
\begin{align}
    \textsc{Replan}(e,w) &:= \frac{1}{N}\sum_{i=1}^N R_i(e,w),\\
    \label{eq:replan}
    \textsc{Len}_{\min}(e,w) &:= \min_{i:\,S_i(e,w)=1} L_i(e,w)\\
    \textsc{Succ}(e,w) &:= \frac{1}{N}\sum_{i=1}^N S_i(e,w).
    \label{eq:success}
\end{align} \label{eq:utility}
\noindent For attempt $i\in\{1,\dots,N\}$, the replanning cost of underspecified instructions is defined as the number of queries $R_i(e,w)$, efficiency is measured by the realized path length $L_i(e,w)$, and $S_i(e,w)\in\{0,1\}$ is an indicator of reaching the goal within the episode budget (i.e., maximum number of steps).
If no attempt succeeds, we set $\textsc{Len}_{\min}(e,w)$ to the episode budget to penalize failures.
These are then combined to define explanation utility as
\begin{align}
    U(e,w) := & -\alpha \, \textsc{Replan}(e,w) -\beta \, \textsc{Len}_{\min}(e,w)  \nonumber \\
              &+\gamma \, \textsc{Succ}(e,w),
              \label{eq:utility}
\end{align}
with $\alpha,\beta,\gamma \ge 0$. Intuitively, effective procedural explanations minimize replanning, support efficient action, and robustly enable goal completion.

\noindent \textbf{Speaker model (RSA-style explanation choice).}
Finally, we treat the explainer as choosing an explanation that maximizes expected utility for a recipient who will use LLM translation and planning.
The pragmatic speaker distribution is
\begin{align}
    S(e \mid w) \propto \exp\left(\lambda\,U(e,w)\right),
\end{align} \label{eq:rsa}
\noindent where $\lambda$ is an inverse temperature parameter controlling how strongly the speaker favors high-utility explanations.

\section{Results}
We first describe the explanation corpus and map structure. We then test whether model-ranked explanation quality predicts helpfulness judgments and navigation performance, and compare the full model to simpler baselines.
\subsection{Descriptions of collected explanations}

\noindent\textbf{Corpus and maps description.} 
The final corpus from Exp.~1 contained $1,200$ explanations across $24$ maps (avg. $21.16 \pm 13.68$ words).
Qualitatively, explanations ranged from value-focused goal descriptions (``top-left corner''), or policy-focused descriptions of either step-by-step action sequences (``go up twice'') or high-level contingencies (``if you see X, do Y'').
In a keyword-based coding (e.g., ``center'' for value, ``steps'' for low-level policy), both types were often mentioned non-exclusively, such that mixed explanations were most common (value: $75.4\%$; low-level policy: $74.7\%$; high-level policy: $96.8\%$; both: $58.3\%$).

To relate explanation style to map structure, we computed graph properties of the reachable state space from the start: (i) shortest-path distance to the goal; (ii) brittleness, measured by the fraction of dead ends (degree $=1$) and corridors (degree $=2$); (iii) and openness, measured by the fraction of emptiness (degree $\geq 3$). 
Across maps ($n=24$), explanation length was unrelated to shortest-path distance (Spearman $\rho=.08$, $p=.69$), but increased strongly with brittleness ($\rho=.73$, $p<.001$), whereby maps with more corridors and dead ends elicited longer explanations. Intuitively, dead ends and narrow corridors make deviations costly, encouraging more explicit, step-by-step guidance. 
Oppositely, maps with more decision points had shorter explanations ($\rho=-.74$, $p<.001$), consistent with the idea that when many alternative routes are available, goal-level guidance suffices because mistakes are easy to recover from.

\noindent\textbf{Stimulus selection for Experiments~2--4.}
To select stimuli with controlled quality differences, we used the open-sourced DeepSeek\-R1 (70B) as the LLM $\mathcal{T}(e,w)$. 
We ranked explanations and defined \emph{Good}, \emph{Medium}, and \emph{Bad} bins using quantiles of $U(e)$ (higher is better).
This gave us a model-based ordering that we tested against human judgments of explanation quality (Exp.~2) and navigation performance (Exp.~3-4).
This resulted in $72$ explanations ($3$ per map), with length $20.98 \pm 14.78$ words.

\subsection{Behavioral analysis}
To evaluate model-ranked explanations, we analyzed both subjective judgments of helpfulness (Exp.~2) and objective navigation performance (Exp.~3-4).
Across analyses, we test whether explanation quality (Eq.~\ref{eq:utility}) corresponded to human judgments and produced measurable improvements in navigation performance under partial observability.

\noindent\textbf{Explanation utility aligns with human judgments.}
We first tested whether model-categorized explanation quality predicted perceived helpfulness. \autoref{fig:regression}b shows averaged ratings by explanation quality. Consistent with preregistered predictions, explanations labeled as \emph{good} were rated as more helpful than \emph{medium} explanations, which in turn were rated higher than \emph{bad} explanations, yielding a clear monotonic ordering (One-sided Welch's t-tests: Good $>$ Medium: $t(49)=13.25, p<.001, d=1.86$; Medium $>$ Bad: $t(49)=18.48, p<.001, d=2.59$).
A Bayesian linear mixed-effects model with ordered quality (Bad$=0$, Medium$=1$, Good$=2$) and random intercepts for participants and maps confirmed this monotonic increase ($\beta=17.87$, 95\% CI [$17.02$, $18.74$]): explanations predicted to help a partially informed listener are consistently judged as more helpful.

\begin{figure}[!t]
    \centering
    \includegraphics[width=0.48\textwidth]{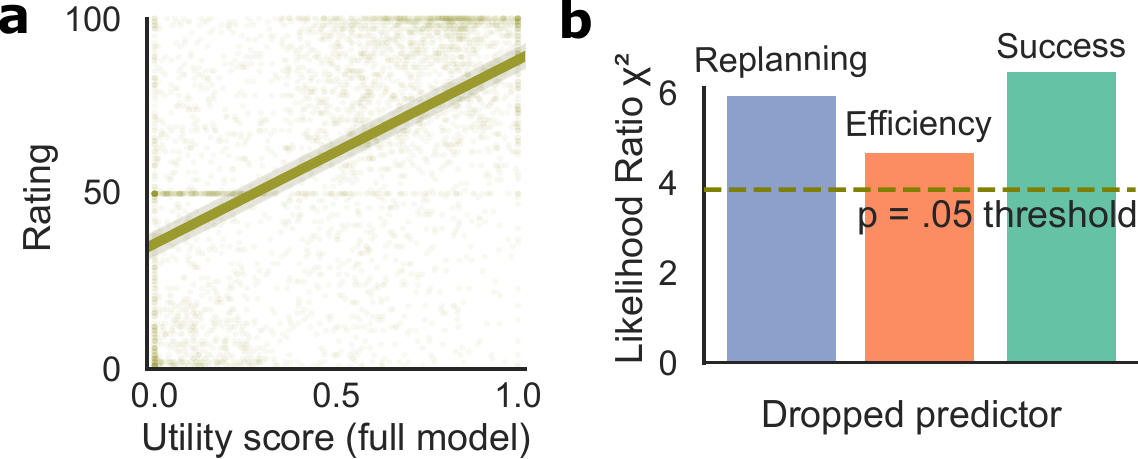}
    \caption{\textbf{Model comparison and component ablations.}
    (a) Points show participants’ helpfulness ratings by full-model utility score; the line shows the predicted–subjective helpfulness regression.
    (b) Leave-one-component-out ablations. Bars show the likelihood-ratio $\chi^2$ statistic versus the full model; larger values indicate greater loss of fit. The dashed line marks the $p=.05$ threshold; bars above it indicate significant loss of fit after predictor removal.}
    \label{fig:baseline}
\end{figure}

\noindent\textbf{Higher-quality explanations produce more efficient navigation.}
We tested whether navigation efficiency varies with explanation quality. Relative to no-explanation trials, providing explanations substantially reduced path length overall ($\beta=-7.73$, 95\% CI [$-9.05$, $-6.12$]), showing that even imperfect guidance helps under partial observability.
Within explanation trials, we fit a Bayesian mixed-effects model with explanation quality entered as an ordinal score (Bad$=0$, Medium$=1$, Good$=2$) and random intercepts for participants and maps. Higher-quality explanations predicted shorter paths ($\beta=-3.65$, 95\% CI [$-4.25$, $-3.00$]).

\subsection{Computational model evaluation}
We next evaluated whether the full utility model predicts human helpfulness judgments better than simpler alternatives. We scored explanations by simulating how a partially informed listener would translate the explanation into executable guidance and then act.

We compared the full utility model to two variants. \textit{Length-only:}
As a simple processing heuristic, we scored explanations by negative word count, $U_{\textsc{Len}}(e) \propto -\,\textsc{Len}(e)$.
\textit{Direct-action (non-program):}
To test whether an explicit program-like intermediate representation is necessary, we prompted the LLM to output the agent's next action directly from the current observation and explanation, without an explicit policy and value representation, and without a replanning loop. We evaluated each explanation over $N$ attempts per map and scored it using success and path length (capped on failure), but \emph{without} replanning cost:
$U_{\textsc{Direct}}(e,w) := \delta\,\textsc{Succ}(e,w) - \alpha\,\mathbb{E}[L(e,w)]$, with fixed $\alpha,\delta \ge 0$ across models.

\noindent\textbf{Analyzing what makes a good explanation.}
For comparability, we normalized each model's score to $[0,1]$ within each map and analyzed only explanations that were rated by participants. 
We then fitted a mixed-effects regression, predicting helpfulness from the full model's utility score, a length-only score, and a direct-action score entered simultaneously, with random intercepts for participants and maps. The utility score uniquely predicted higher helpfulness, $\beta_{\textsc{Util}}=38.47$, 95\% CI$ =[36.24, 40.17]$. In contrast, the length-only and direct-action predictors were weaker ($\beta_{\textsc{Len}}=.85$, 95\% CI$ =[-1.23, 2.94]$, $\beta_{\textsc{Direct}}=.08$, 95\% CI$ =[-2.10, 2.28]$). \autoref{fig:baseline}a shows helpfulness as a function of utility, with the partial regression line from the joint model holding the other predictors at their means.
Overall, helpfulness reflects more than brevity: participants preferred explanations that the utility model predicts will support reliable and efficient guidance. Leave-one-component-out ablations further showed that all three utility terms contributed to predicting judgments (\autoref{fig:baseline}b). Removing \textsc{Replan} or \textsc{Succ} produced the largest losses, suggesting that people penalize explanations requiring frequent reinterpretation or yielding unreliable guidance; removing path efficiency had a smaller but reliable effect.

\begin{figure}[t]
    \centering
    \includegraphics[width=0.48\textwidth]{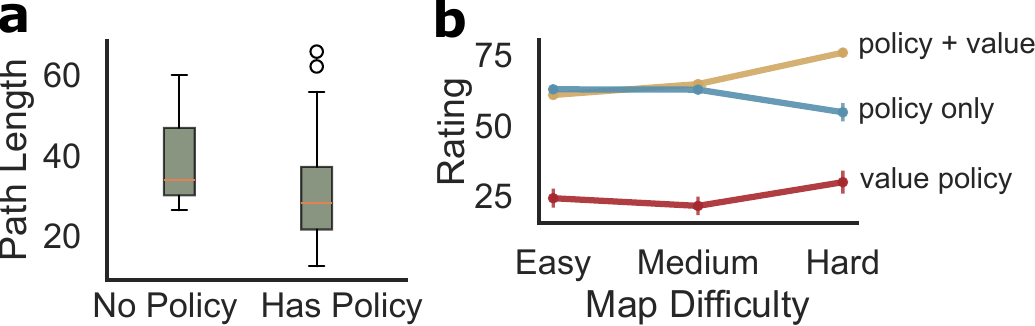}
    \caption{\textbf{Policy vs.\ value content in explanations.}
    (a) Navigation path length for explanations that contain policy guidance vs.\ those that do not.
    (b) Mean helpfulness ratings by map difficulty for explanations containing policy+value information, policy-only information, or value-only information. Error bars show standard errors.}
    \label{fig:decompose}
    \vspace{-1.5em}
\end{figure}

\noindent\textbf{Linguistic strategies and structure.}
We analyzed rated explanations to characterize strategies associated with higher utility. We coded whether an explanation primarily provided a \emph{policy} (procedural actions), a \emph{value} signal (goal- or landmark-oriented direction), or both.
Policy information improved navigation efficiency: trials with policy content yielded shorter paths than trials without policy content (\autoref{fig:decompose}a; $25.09 \pm 7.36$ vs.\ $33.83 \pm 12.52$; $t=3.20$, $p<.005$, $d=.851$). 
However, helpfulness depended on combining strategies. Explanations integrating policy and value cues were rated highest overall (policy+value: $64.23 \pm 31.50$; policy-only: $62.47 \pm 30.41$; value-only: $24.99 \pm 25.80$), although the difference between policy+value and policy-only was not significant ($t=1.614$, $p=.106$). Importantly, the advantage of combining policy and value increased with map difficulty (\autoref{fig:decompose}b): policy-only guidance degraded on hard maps, whereas policy+value remained robust. 
Overall, these analyses provide converging evidence that our model's notion of communicative utility---grounded in executable guidance for a partially informed listener---captures both the \emph{behavioral} and \emph{linguistic} signatures of what people consider to be good explanations of how to do something.

\begin{table}[t]
\caption{\textbf{Failure modes of LLM translation.} Counts ($n$), mean LLM success rate (Succ.), mean word length (Len.), and mean number of direction words (Dir.) per explanation. Categories are non-exclusive.}
\label{tab:failure}
\centering
\small
\setlength{\tabcolsep}{4pt}
\begin{tabular}{lrrrr}
\toprule
\textbf{Failure mode} & \textbf{$n$} & \textbf{Succ.} & \textbf{Len.} & \textbf{Dir.} \\
\midrule
Direction overload   & 131 & .46 & 27.4 & 4.8 \\
Overcomplicated      & 151 & .40 & 38.1 & 3.2 \\
Overly compressed    &  61 & .19 &  5.9 & 0.6 \\
Spatial ambiguity    &  10 & .13 & 14.2 & 0.4 \\
\bottomrule
\end{tabular}
\vspace{-1em}
\end{table}

\noindent\textbf{Failure modes of LLM translation.}
We next characterized \emph{why} certain explanations fail. 
Coding explanations by length, direction-word density, plan failure rate, and lexical cues produced four non-exclusive categories (\autoref{tab:failure}). 
\emph{Direction overload} consists of turn-by-turn directives (``go up 3, then right 2, then down 4'') that destabilize the planner---a single misaligned step cascades.
\emph{Overcomplicated} explanations are long multi-clause instructions whose conditional structure (``if you see X, then Y, unless Z'') exceeds what the LLM compiler can reliably parse. 
\emph{Overly compressed} explanations which omit needed details and induce uninformed search.  
\emph{Spatial ambiguity} names a region without a referable landmark (``somewhere near the middle''), forcing the planner into repeated re-queries. 
The hardest map (Map~10, $4.7\%$ success) was dominated by direction overload, as long corridors amplified the cost of a single misaligned step. More broadly, the two most frequent failure modes fall at opposite ends of length and directive density, suggesting that failure reflects not simply being too terse or too verbose, but a mismatch between explanation form and actionable guidance.

Human judgments and LLM translation success can also diverge. Using a top-quartile cutoff ($\geq 0.7$ LLM success; $\geq 70$ helpfulness), $19$ explanations are \emph{LLM-only} and $15$ are \emph{human-only}. \emph{Human-only} cases are more than twice as long ($35.1$ vs.\ $16.8$ words), use more direction words ($5.1$ vs.\ $2.5$), and more often include explicit step counts ($60\%$ vs.\ $16\%$): the same surface features humans read as conscientious guidance overload the translator with brittle commitments. \emph{LLM-only} explanations are terse, landmark-anchored statements that compile cleanly but read as under-informative. This double dissociation suggests that bridging the gap requires translators that recover intent from verbose human directives, not speakers writing for the machine.

\section{Discussion}
Procedural explanations are designed to \emph{produce competent action} in a listener who lacks full information, not merely describe the world. In a partially observable navigation task, explanations ranked by our utility model received higher helpfulness ratings and yielded shorter paths than no-explanation baselines, with monotonic gains from Bad to Medium to Good---supporting a behavioral criterion for explanation quality: an explanation is good insofar as it grounds into \emph{executable guidance} that supports planning under uncertainty.
Our framework builds on the RSA intuition that utterances maximize communicative utility, but shifts the target from information gain to downstream behavior: we evaluate an explanation by simulating a partially informed agent that stochastically translates language into guidance for planning and navigation. This predicts that models capturing the translation-and-planning bottleneck should match human judgments better than surface heuristics or planning-free baselines---which our results support. The full model outperformed both a length-only heuristic (so participants do not simply prefer shorter explanations) and a direct-action baseline that skips program-like guidance (so modeling \emph{how} explanations shape sequences of decisions matters).

Our linguistic analyses clarify what executable guidance looks like. Purely procedural, step-by-step explanations can improve efficiency when they remain aligned with the listener's local context, but are brittle under partial observability. By contrast, value- or landmark-oriented information helps the listener re-localize and replan when things go awry. Consistent with this account, explanations integrating procedural steps with goal structure were rated highest overall, and yield larger benefits with increasing map difficulty. This supports an action-under-uncertainty view of instruction giving: good explanations balance specificity (to reduce deliberation) with stable reference points (to avoid replanning).

Several limitations remain. First, we used simplified grid worlds to provide a controlled testbed. However, real-world instructions involve hierarchical goals \citep{colas2025language} and interactive replanning \citep{tomlin2025characterizing}. Second, our listener model is largely static. Aside from partial observability, its physical capabilities (what actions are feasible) and knowledge (what landmarks or conventions are shared) do not change. In natural settings, effective explainers tailor guidance by inferring what a listener can do and what they already know \citep{velez2023teachers}. A natural extension is to model listener knowledge as a latent state inferred online from behavior \citep{zhou2024predictive}, enabling explanations that adapt to individual capabilities in more open-ended environments.

Overall, the results support a view of procedural explanation as utility-guided communication grounded in action: explanations are effective when they reliably translate into plans that work for a partially informed listener, minimize replanning, and support efficient goal achievement.

\clearpage 
\section{Acknowledgments}
We thank Robert Hawkins and Lio Wong for helpful discussions. 
The authors thank the International Max Planck Research School for Intelligent Systems (IMPRS-IS) for supporting HZ. 
HZ and CMW are supported by the European Research Council (ERC) under the European Union’s Horizon 2020 research and innovation programme ($C^4$: 101164709), the Hessian research funding programme LOEWE/4b//519/05/01.002(0022)/119, the Deutsche Forschungsgemeinschaft (German Research Foundation, DFG) under Germany’s Excellence Strategy (EXC 3066/1 ``The Adaptive Mind'', Project No. 533717223), and the Excellence Cluster ``Reasonable AI'' by the Deutsche Forschungsgemeinschaft (German Research Foundation, DFG) under Germany’s Excellence Strategy – EXC-3057. 
TG was supported by grants from the Stanford Institute for Human-Centered Artificial Intelligence (HAI) and from the Cooperative AI Foundation.  BB was supported by the Studienstiftung des deutschen Volkes (German Academic Scholarship Foundation) for her research stay at Stanford.

\printbibliography 
\end{document}